\documentclass[conference,a4paper]{IEEEtran}
\IEEEoverridecommandlockouts

\usepackage[hidelinks]{hyperref}
\usepackage[cmex10]{amsmath}
\usepackage{amssymb,amsfonts}
\usepackage{dblfloatfix}

\usepackage[ruled,vlined]{algorithm2e}
\usepackage{graphicx}
\graphicspath{{Figures/PDF/}{Figures/PNG/}}

\usepackage{booktabs}
\usepackage{siunitx}
\usepackage[numbers,compress]{natbib}
\usepackage{texnames}
\usepackage{bm,bbm}
\usepackage{orcidlink}
\usepackage{tabularx}
\usepackage{multirow}
\usepackage{multicol}
\usepackage{subfig}
\usepackage{caption} 
\usepackage{tikz}
\usepackage{xspace}
\usepackage[acronym, shortcuts]{glossaries}
\usepackage{colortbl}
\usepackage{array}

\newcommand{\best}[1]{\pdfliteral direct {2 Tr 0.3 w}#1\pdfliteral direct {0 Tr 0 w}} 
\newcommand{\secondbest}[1]{{\underline{{#1}}}}

\newcolumntype{L}{>{\raggedright\arraybackslash}X}
\newcolumntype{C}{>{\centering\arraybackslash}X}
\newcolumntype{R}{>{\raggedleft\arraybackslash}X}

\newacronym{eo}{EO}{Earth Observation}
\newacronym{sits}{SITS}{Satellite Image Time Series}
\newacronym{ts}{TS}{Time Series}
\newacronym{xai}{XAI}{eXplainability Artificial Intelligence}
\newacronym{cropharvestB}{{CropH-b}}{CropHarvest binary}
\newacronym{cropharvestM}{{CropH-m}}{CropHarvest multi}
\newacronym{treesat}{{TreeSatAI-TS}}{TreeSatAI-Time-Series}

\begin{document}

\title{On What Depends the Robustness of Multi-source Models to Missing Data in Earth Observation?
\thanks{F. Mena acknowledges the financial support from the chair of Prof. A. Dengel with RPTU. Corresponding e-mail: f.menat@rptu.de.}
}

\author{\IEEEauthorblockN{Francisco Mena$^{1,2}$\orcidlink{0000-0002-5004-6571},
Diego Arenas$^{2}$\orcidlink{0000-0001-7829-6102}, Miro Miranda$^{1,2}$\orcidlink{0009-0002-8195-9776}, and Andreas Dengel$^{1,2}$\orcidlink{0000-0002-6100-8255}}
\IEEEauthorblockA{
$^{1}$ University of Kaiserslautern-Landau (RPTU), Kaiserslautern, Germany \\
$^{2}$ German Research Center for Artificial Intelligence (DFKI), Kaiserslautern, Germany 
}
}

\maketitle
\begin{abstract}
In recent years, the development of robust multi-source models has emerged in the Earth Observation (EO) field. These are models that leverage data from diverse sources to improve predictive accuracy when there is missing data. Despite these advancements, the factors influencing the varying effectiveness of such models remain poorly understood. 
In this study, we evaluate the predictive performance of six state-of-the-art multi-source models in predicting scenarios where either a single data source is missing or only a single source is available. 
Our analysis reveals that the efficacy of these models is intricately tied to the nature of the task, the complementarity among data sources, and the model design. 
Surprisingly, we observe instances where the removal of certain data sources leads to improved predictive performance, challenging the assumption that incorporating all available data is always beneficial. 
These findings prompt critical reflections on model complexity and the necessity of all collected data sources, potentially shaping the way for more streamlined approaches in EO applications.
\end{abstract}

\begin{IEEEkeywords}
	Missing data, Earth observation, Deep learning, Multi-source learning, Robustness.
\end{IEEEkeywords}

\section{Introduction}

Various solutions in the \gls{eo} field rely on multi-source data. The diversity and availability of numerous data sources has allowed researchers to build more complex and effective models for different applications \cite{camps2021deep,mena2024common}. 
However, the use of multiple data sources in model design leads to dependence on the availability of all of them for prediction.
Thus, the literature has shown that the lack of data produces a negative impact on predictive performance \cite{hong2021more,saintefaregarnot2022multi,mena2024igarss}. This effect can vary significantly depending on \textit{unknown} factors.

Although modern instruments are used to collect data about the Earth's surface, missing data is an inherent phenomenon in Earth observation \cite{shen2015missing}.
As this collection occurs in real-world environments, different situations may affect the availability of data.
One example is cloud occlusion, affecting the spatio-temporal availability of optical images.
Another case corresponds to sensor failures and errors, just like the Landsat 7 ETM+ SLC-off problem after 2003, or the Sentinel-1b satellite that stopped operating at the end of 2021.

Currently, multiple studies in the \gls{eo} field have introduced robust multi-source models to missing data. For instance, by simulating missing data during training with dropout techniques at spectral \cite{haut2019hyperspectral}, spatial \cite{fasnacht2020robust}, temporal \cite{saintefaregarnot2022multi}, or sensor \cite{mena2024notyet} dimensions, or by forcing similarities between the data sources \cite{kampffmeyer2018urban,xie2023co,mena2024notyet,guo2024skysense}.
However, none of them has assessed in what scenarios and what are the characteristics that make some multi-source models work better than others. 

In our work, we evaluate six multi-source models in different source-wise missing data scenarios. We use \gls{eo} datasets that include temporal and static (single-date) data sources in three classification tasks.
We train the models in a full-data scenario and validate them when data sources are missing.
We observe that the robustness of multi-source models to missing data depends on the predictive task, the complementarity between the data sources, and the model design.

Inspired by a data-centric focus \cite{roscher2024better}, we explore questions such as \textit{are additional data sources beneficial in the modeling?} \textit{does the individual performance of the data sources relate to their complementary effectiveness in the task?} and \textit{is a robust model effective for all cases of source-wise missing data?}

\section{Related work}

Some works in the literature have explored the effect of missing data in multi-source models.
For instance, Hong et al. \cite{hong2021more} analyze how the predictions deteriorate in two-source models with different fusion strategies. Similarly, Mena et al. \cite{mena2024igarss} evaluate different methods to handle source-wise missing data.
They obtain that the missing of the optical data source affects in a greater extent than others.
In Ekim et al. \cite{ekim2024deep} simulate missing data is used to compute a proxy to importance scores, similar to perturbation methods in \gls{xai}. 
They found that the data sources with higher individual performance are not necessarily the important ones when they are missing.
Additionally, Chen et al. \cite{chen2024novel} and Gawlikowski et al. \cite{gawlikowski2023handling} assess the robustness of their proposal in single-date data sources.
Instead, in our study, we evaluate various robust multi-source models to discover the common patterns.
Our study involves evaluating different fusion strategies, robustness components, and missing data scenarios with static and temporal data sources.

\section{Missing data in the Earth observation field}

\begin{figure*}[!t]
    \centering
    \includegraphics[width=0.94\linewidth]{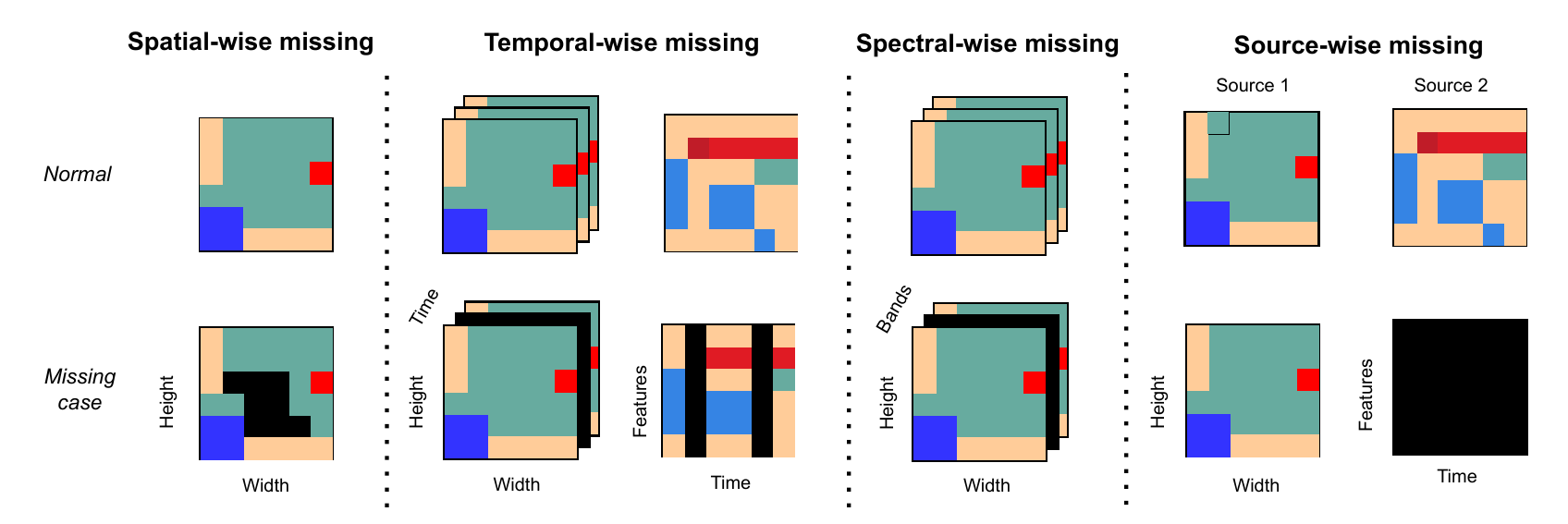}
    \caption{Types of missing data in the \gls{eo} field. For the temporal missing, two cases are shown (spatial and feature wise).}\label{fig:missing}
\end{figure*}
There are different types of missing data, as illustrated in Fig.~\ref{fig:missing}. The spatial-wise is when a portion of the image is missing in spatial data, while the temporal-wise corresponds to missing time-steps in temporal data.
In addition, the spectral-wise indicates that a band or spectral channel is missing in multi-spectral data. 
Lastly, when all features of a data source are unavailable, exemplifies the extreme case of source-wise missing data. 
In our work, we focus on the latter case when it occurs during inference in multi-source learning models.

\subsection{How to handle source-wise missing data?}

In order to use the trained multi-source model in source-wise missing data scenarios, an intervention is required, usually involving a data processing step.

One data processing option is to impute the missing features with the same numerical value for all, such as with zero \cite{hong2021more} or the training-set-average \cite{mena2024igarss}. 
However, the missing features can also be replaced with existing values observed in the training set, e.g. by searching for a similar sample with the available data sources \cite{srivastava2019understanding}. 
A third case corresponds to dropping the part of the model that depends on the data source that is missing and fine-tune the rest of the model, as is the case of geospatial foundational models \cite{guo2024skysense,astruc2025omnisat}. 
At last, multi-source models can be designed to ignore the missing data by using an adaptive aggregation, such as averaging in ensembles \cite{srivastava2020fine,mena2024notyet}.

\subsection{How to be robust to source-wise missing data?}

Some components can be included in the model design to reduce the decline in predictive performance when there is missing data. This leads to robust model proposals.

One research line is based on simulating missing data during training, so the model can learn the missing patterns that will be expected, e.g. during inference. 
An alternative is to do it randomly. For instance, using a sensor dropout technique that replaces all features of random sensors with zeros \cite{mena2024notyet} or by masking out the features of random sensors in attention-based models \cite{chen2024novel,astruc2025omnisat}.
Another option is to simulate source-wise missing cases. For instance, Gawlikowski et al. \cite{gawlikowski2023handling} use this by replacing the features of the missing data source with zeros.

Another research line is based on enhancing the similarity learning between data sources. For instance, by sharing weights in per data source layers \cite{mena2024notyet,zheng2021deep}, i.e. a weight-based similarity enhancement. 
Similarly, by forcing features (hidden dimensions \cite{wang2024decoupling,ienco2024discom}, or decisions \cite{xie2023co}) in per data source models to be similar, i.e. a data-based similarity enhancement.
Lastly, the cross-source similarity can be enhanced by reconstructing a data source based on another one \cite{nedungadi2024mmearth,chen2024novel}.

\section{Evaluation}

\subsection{Datasets}
We use three pixel/patch-wise classification datasets with static (single-date) and temporal data sources. 

\paragraph{\gls{cropharvestB}}
We use the CropHarvest dataset for crop recognition with four data sources \cite{tseng2021crop}.
This involves a cropland (binary) classification at a specific region (re-sampled at 10x10 $m^2$).
The dataset has 69,800 samples around the globe between 2016 and 2021. 
Each sample has three temporal data sources at $10m$ resolution: multi-spectral optical \gls{sits} (11 bands), radar \gls{sits} (2 polarization bands), and weather \acrshort{ts} (2 bands).
These \gls{ts} have one value per month over 1 year.
Besides, the samples have one static data source, the topographic information (2 bands).

Furthermore, we use a multi-class version of this data, \gls{cropharvestM}. 
This is a subset of 29,642 samples with a crop-type classification task. 
It includes 10 crop types (mutually exclusive) and the same four input sources available in \gls{cropharvestB}. 
Since there is no test partition in this dataset, we use a standard 10-fold cross validation for this.

\paragraph{\gls{treesat}}
We use the TreeSatAI dataset \cite{ahlswede2022treesatai} extended with temporal data sources in \cite{astruc2025omnisat}.
This involves a multi-label classification of tree species in a specific region (60x60 $m^2$).
There are 38,520/6,810/5,044 samples for training, validation, and testing respectively, collected between 2017 and 2020 in Germany. 
Each sample has two temporal data sources at a $10m$ resolution: multi-spectral optical \gls{sits} (12 bands) and radar \gls{sits} (2 polarization bands).
These \gls{ts} have around 10 images per month over 1 year.
Another source is the single-date aerial image (four bands: RGB + infrared band) with a high spatial resolution of $0.2m$.

\begin{table*}[!t]
	\centering
	\caption{Weighted F1 score in different cases of source-wise missing data in the \textbf{\gls{cropharvestB}} data. The results of the best method in each missing case is in bold, and the second best is underlined. In parentheses is the complementarity score when a single source is missing. We use the symbol $\dagger$ when this score is negative.}\label{tab:cropbinary}
	\begin{tabularx}{0.975\textwidth}{cccc|c|LLLLLL}
		\toprule
         \multicolumn{4}{c|}{\textbf{Data sources}} & \textbf{Single} & \textbf{ISensD} & \textbf{FSensD} & \textbf{OOD-f} & \textbf{TIMML} & \textbf{ESensI}  &  \textbf{Co-Ens} \\ 
         Optical  & Radar & Weather & Topography & \\
        \midrule
        $\checkmark$ & $\checkmark$ & $\checkmark$ & $\checkmark$  &   &  	$80.0$ &	$82.1$ &	$\best{84.7}$  &	$\secondbest{83.1}$ &	$81.4$ & $79.2$ \\
        \midrule
         - & $\checkmark$ & $\checkmark$ & $\checkmark$ & & $72.6$ $(10.2)$ & $79.0$ $(3.9)$ &	$\best{82.5}$ $(2.7)$ &	$\secondbest{80.1}$ $(3.7)$ &  $78.1$ $(4.2)$ &	$76.8$ $(3.1)$  \\
        $\checkmark$ & - & $\checkmark$ & $\checkmark$ & & $79.4$ $\;(0.8)$ & $81.8$  $(0.4)$ &	$\best{84.2}$ $(0.6)$ &	$\secondbest{82.8}$ $(0.4)$ &	$82.4$ $\;(\dagger)$  & $79.3$ $\;(\dagger)$  \\
        $\checkmark$ & $\checkmark$ & - & $\checkmark$ & & $\secondbest{80.4}$ $\;(\dagger)$ & $80.0$ $(2.6)$ &	$\best{81.3}$ $(4.2)$ &	$\best{81.3}$ $(2.2)$ &	$79.7$ $(2.2)$  & $78.4$ $(1.0)$ \\
        $\checkmark$ & $\checkmark$ & $\checkmark$ & - &  & $80.0$ $\;(0.0)$ & $82.1$ $(0.0)$ &	$\best{84.2}$ $(0.5)$ &	$\secondbest{83.1}$ $(0.0)$ &	$81.2$ $(0.3)$ & $80.0$ $\;(\dagger)$ \\
        \midrule
		$\checkmark$ & - & - & - & $81.8$ &  ${80.0}$ &	$79.7$ &	$79.5$ &	$\best{80.9}$ &	$\secondbest{80.8}$  & $\best{80.9}$  \\
         - & $\checkmark$ & -  & - &  $71.4$ & $44.6$ &	$69.5$ &	${69.6}$  &	${70.1}$ &	$\secondbest{70.7}$ &	$\best{72.2}$ \\
         - & -  & $\checkmark$ & - & $77.8$ &    $60.8$ &	$75.8$ &	$\secondbest{77.2}$ &	 $76.2$	 &	$\best{77.3}$ &  $\secondbest{77.2}$ \\
         - & - & -  & $\checkmark$ & $67.7$ &  & $56.5$ &	$\secondbest{66.9}$ &	$63.8$ &	$65.9$ & $\best{68.3}$ \\
         \midrule
         \multicolumn{4}{c|}{Average} & $74.7$ & $72.2$ & $77.3$ & $\best{78.9}$ & $\secondbest{77.9}$ & $77.6$ & $76.9$ \\
        \bottomrule
	\end{tabularx}
\end{table*}

\begin{table*}[!t]
	\centering
	\caption{Weighted F1 score in different cases of source-wise missing data in the \textbf{\gls{cropharvestM}} data. The results of the best method in each missing case is in bold, and the second best is underlined. In parentheses is the complementarity score when a single source is missing. We use the symbol $\dagger$ when this score is negative.}\label{tab:cropmulti}
	\begin{tabularx}{\textwidth}{cccc|c|LLLLLL}
		\toprule
         \multicolumn{4}{c|}{\textbf{Data sources}} & \textbf{Single} & \textbf{ISensD} & \textbf{FSensD} & \textbf{OOD-f} & \textbf{TIMML} & \textbf{ESensI}  & \textbf{Co-Ens} \\ 
         Optical & Radar & Weather & Topography & \\
        \midrule
        $\checkmark$ & $\checkmark$ & $\checkmark$ & $\checkmark$  &   & $71.3$ &	$69.5$ &	$\best{73.2}$ &	$\secondbest{72.8}$ & $71.2$ &	$63.6$ \\
        \midrule
         - & $\checkmark$ & $\checkmark$ & $\checkmark$ & & $46.1$ $(54.7)$ &	${60.8}$ $(14.3)$ &	$\best{65.2}$ $(12.3)$ &	$\secondbest{63.3}$ $(15.0)$ &	$59.5$ $(19.8)$ & $55.5$ $(14.6)$ \\
        $\checkmark$ & - & $\checkmark$ & $\checkmark$ & & $69.8$ $\;(2.1)$ &	$68.1$ $\;(2.0)$ &	$69.9$ $\;(4.7)$ &	$\best{71.4}$ $\;(2.0)$ & $\secondbest{70.2}$ $\;(1.5)$ & $60.0$ $\;(6.0)$	 \\
        $\checkmark$ & $\checkmark$ & - & $\checkmark$ & &  $70.5$ $\;(1.1)$ &	$69.1$ $\;(0.7)$ &	${72.0}$ $\;(1.7)$  &	$\secondbest{73.0}$ $\;(\dagger)$ & $\best{73.1}$ $\;(\dagger)$ & $66.3$ $\;(\dagger)$	 \\
        $\checkmark$ & $\checkmark$ & $\checkmark$ & - &  & $71.3$ $\;(0.0)$ &	$69.5$ $\;(0.0)$ &	$\best{73.1}$ $\;(0.1)$ &	$\secondbest{72.8}$ $\;(0.0)$ & $71.2$ $\;(0.1)$ &	  $65.5$ $\;(\dagger)$ \\
        \midrule
		$\checkmark$ & - & - & -  & $72.8$ &  ${69.1}$ &	$67.6$ &	$69.3$  &	$\secondbest{71.8}$ &	$\best{72.5}$ & $70.8$ \\
         - & $\checkmark$ & - & - &  $55.8$   &  $17.4$ &	$46.9$ &	${51.6}$ &	$53.7$ & $\secondbest{54.8}$ & $\best{56.5}$	 \\
         - & -   & $\checkmark$ & -  & $46.9$  &  $15.7$ &	$42.0$ &	$42.8$ &	$44.8$ & $\secondbest{45.0}$ &	$\best{47.4}$  \\
         - & - &-  & $\checkmark$ & $28.0$ &   & $\secondbest{15.3}$ &	$15.1$ & $\; 3.1$ & $14.8$ &	$\best{30.0}$  \\
         \midrule
         \multicolumn{4}{c|}{Average} & $50.9$ & $53.9$ &  $56.5$ & $\best{62.5}$ & ${58.5}$ & $\secondbest{59.1}$ & $57.3$  \\
        \bottomrule
	\end{tabularx}
\end{table*}

\subsection{Compared multi-source models}
We consider the four models from the \gls{eo} field that simulate missing data sources. 
\textbf{ISensD} \cite{mena2024notyet}, an input-level fusion model that uses sensor dropout by replacing missing features with zero.
\textbf{FSensD}, a feature-level fusion model with cross-attention that uses sensor dropout by masking out the missing features of the attention. This method is adapted from \cite{chen2024novel,astruc2025omnisat} to supervised learning, without pre-training.
\textbf{OOD-f} \cite{gawlikowski2023handling}, a feature-level fusion model with concatenation that simulates all cases of source-wise missing data, which we extended from a two-source formulation.
\textbf{TIMML} \cite{xu2024transformer}, a feature-level fusion model with cross-attention that uses sensor dropout and auxiliary prediction losses per data source.
We also consider two ensemble-based models that enhance similarity between data sources.
\textbf{ESensI} \cite{mena2024notyet}, a model that uses shared prediction heads in the per-source layers.
\textbf{Co-Ens} \cite{xie2023co}, a model using a similarity constrain in the per-source prediction.

We run each model five times and report the average results.

\subsection{Implementation}
We use standard encoders for each data source.
For all temporal data sources in the datasets, we use TempCNN \cite{pelletier2019temporal}, a 1D convolutional network applied over time.
For the pixel-wise topographic information (CropHarvest data) we use a standard MLP, while for the aerial image (TreeSatAI-TS data) we use ResNet-50 \cite{he2016deep}, a 2D convolutional neural network with skip connections. 
For all encoders (apart from ResNet) we use two hidden layers and a final embedding layer of 128 units. 
In addition, we use a prediction head with a hidden layer of 128 units and a final linear layer for prediction.
We use the optimizer setting of the original proposal in each model. We tried to normalize by using the same optimizer, but the results from each method were worse.
For a fair comparison, we train over 100 epochs with a batch size of 128 and an early stopping criterion with patience of 5.

\subsection{Results}

\begin{table*}[!t]
	\centering
    \caption{Weighted F1 score  in different cases of source-wise missing data in the \textbf{\gls{treesat}} data.  The results of the best method in each missing case is in bold, and the second best is underlined. In parentheses is the complementarity score when a single source is missing. We use the symbol $\dagger$ when this score is negative.}\label{tab:treesat}
	\begin{tabularx}{0.8\textwidth}{ccc|c|LLLL}
		\toprule
         \multicolumn{3}{c|}{\textbf{Data sources}} & \textbf{Single} & \textbf{FSensD} & \textbf{OOD-f} & \textbf{TIMML} & \textbf{ESensI}  \\ 
         Optical & Radar & Aerial & \\
        \midrule
        $\checkmark$ & $\checkmark$ & $\checkmark$  &   & {$61.8$}  & {$\best{68.0}$} & $\secondbest{65.6}$ & {$62.6$}   \\
        \midrule
        - & $\checkmark$ & $\checkmark$ & & $59.3$ $(4.1)$  & $\best{65.3}$ $\;(4.2)$ & $\secondbest{64.7}$ $\;(1.4)$ & $63.0$  $\;(\dagger)$  \\
        $\checkmark$ & - & $\checkmark$ & & $61.1$ $(1.1)$& $\best{66.1}$ $\;(3.0)$ & $65.2$ $\;(0.7)$ & $\secondbest{65.6}$  $\;(\dagger)$    \\
        $\checkmark$ & $\checkmark$ & -  & &  $\secondbest{57.5}$ $(7.4)$ & $27.1$  $(150.8)$ & $\best{59.3}$ $(10.7)$ & ${56.4}$ $(10.9)$   \\
        \midrule
		$\checkmark$ &- &- & $65.8$ & ${53.5}$  & $\;5.5$ & $\best{55.5}$ & $\secondbest{55.3}$  \\
         -& $\checkmark$ & - &  $55.0$ & ${51.3}$   & $10.4$   & $\best{52.6}$ & $\secondbest{52.0}$  \\
         -& - & $\checkmark$ & $63.7$  & $56.0$   & $\secondbest{62.2}$  & $\best{63.3}$ &  $\secondbest{62.2}$  \\
         \midrule
         \multicolumn{3}{c|}{Average} & $61.5$ & ${57.2}$ & $43.5$ & $\best{60.9}$ & $\secondbest{59.6}$  \\
        \bottomrule
	\end{tabularx}
\end{table*}

We show the results in the binary and multi-class crop classification in Table~\ref{tab:cropbinary} and \ref{tab:cropmulti} respectively. We display the effect when a single data source is missing as well as when a single data source is available.
The decline in performance when data source $s$ is missing or is the only one available depends on the usefulness of $s$ for the predictive task, reflected in the single predictive performance. For instance, the weather data source in \gls{cropharvestB} and the radar source in \gls{cropharvestM} data.
Furthermore, when a single data source is available for prediction (like the optical), the drop in performance is considerable high, as observed in the literature \cite{hong2021more,mena2024notyet}.
Overall, the OOD-f and TIMML models are the most robust when a single data source is missing, while when a single data source is available, the EsensI and Co-Ens are.

In addition, we include the complementary score as the relative drop in performance when a single data source is missing. This score reflects how useful this data source is for prediction, similar to perturbation methods in \gls{xai}. This score, shown in parenthesis in both Tables ~\ref{tab:cropbinary} and \ref{tab:cropmulti}, corresponds to $(P_A - P_{m(s)})/P_{m(s)}$, with $P_A$ the prediction with all data sources, and $P_{m(s)}$ the prediction when data source $s$ is missing.
As expected, the highest complementarity score is related to the optical \gls{sits} for all multi-source models, except with the weather \gls{ts} in \gls{cropharvestB} for the OOD-f model.

For the tree-type classification, we show the weighted F1 scores and complementarity scores in Table~\ref{tab:treesat}.
We notice similar results than for the crop classification tasks, except that the highest complementarity score for all models is associated with the aerial data source and its single-date high spatial resolution information.
We did not include ISensD and Co-Ens as these models are incompatible with heterogeneous spatio-temporal data sources and multi-label task, respectively.

\section{Discussion}

Overall, we notice that the robustness of multi-source models to missing data depends on different factors.

\begin{enumerate}
    \item How informative the data source is on the predictive task, related to the single predictive performance.
For instance, in crop classification, the weather \gls{ts} is more effective in the binary classification than the multi-class, and the radar \gls{sits} is more effective in the multi-class than the binary classification.
This could be explained as crops are expected to grow under specific climate conditions, but it is difficult to distinguish which is the crop-type that is growing in that climate. On the other hand, surface characteristics (via radar) can be used to better distinguish the crop-type.
\item How complementary the data source is regarding the others, based on the complementarity score. 
For instance, the aerial data source complements the optical and radar \gls{sits} to a greater extent in the tree-type classification, even though it does not offer the best single predictive performance.
\item How invariant the model design is to ignore missing features.
For instance, the OOD-f model is more robust to missing a single data source, while Co-Ens model is more robust when a single data source is available.
\end{enumerate}

In addition, the following research questions are discussed.

\paragraph{Does the individual-source performance relate to their complementary effectiveness in the task?}
Not necessarily.
For instance, using the OOD-f model, the weather \gls{ts} has a higher complementarity score than the optical \gls{sits} in the \gls{cropharvestB} data, despite having a lower single predictive performance. 
This effect mainly depends on the information and complementarity between the data sources.

\paragraph{Are additional data sources beneficial in the modeling?}
Not necessarily, as in some models the predictions with less data sources outperform the ones with full-data.
For instance, using the radar \gls{sits} in the \gls{treesat} data 
with ESensI and Co-Ens models. Besides, the weather \gls{ts} decreases the predictions of TIMML, ESensI, and Co-Ens models in the \gls{cropharvestM} data.
Thus, the benefit of additional data sources depends on the factors described before: the predictive task, the complementarity between data sources, and model design.

\paragraph{Is a robust model effective for all cases of source-wise missing data?}
We notice the opposite, that the results vary greatly depending on which missing data scenario  occurs. 
For instance, when a single data source is missing in the \gls{cropharvestM} data, the best results varies between OOD-f, TIMML, and ESensI models. 
We notice that for single-source predictions, the models are unreliable and very data-dependent.
Then, if during deployment a single data source is expected, it is better to train a model specialized for this scenario.

\section{Conclusion}
We present a study evaluating different multi-source models to detect the common factors that affect the effectiveness of the robustness. 
We find a structure in the results with three main factors of what the robustness depends on, the predictive task, the complementarity between data sources, and the model design. 
Furthermore, we notice multi-source models that are more robust when a single data source is missing (OOD-f and TIMML), or when a single data source is available for prediction (ESensI and Co-Ens).
Lastly, we realize that less data sources is better in some cases, which may motivate future work on careful data source selection based on missing data analysis in the \textit{same} trained model.
This relates to perturbation methods in \gls{xai}, such as its usage in feature selection \cite{najjar2024data} or quantifying modality contribution \cite{ekim2024deep}.

\small
\bibliographystyle{IEEEtranN}
\bibliography{main}

\end{document}